# An Approach to Reducing Annotation Costs for BioNLP

**Michael Bloodgood**
Computer and Information Sciences
University of Delaware
Newark, DE 19716
bloodgoo@cis.udel.edu

**K. Vijay-Shanker**
Computer and Information Sciences
University of Delaware
Newark, DE 19716
vijay@cis.udel.edu

## 1 Introduction

There is a broad range of BioNLP tasks for which active learning (AL) can significantly reduce annotation costs and a specific AL algorithm we have developed is particularly effective in reducing annotation costs for these tasks. We have previously developed an AL algorithm called *ClosestInitPA* that works best with tasks that have the following characteristics: redundancy in training material, burdensome annotation costs, Support Vector Machines (SVMs) work well for the task, and imbalanced datasets (i.e. when set up as a binary classification problem, one class is substantially rarer than the other). Many BioNLP tasks have these characteristics and thus our AL algorithm is a natural approach to apply to BioNLP tasks.

## 2 Active Learning Algorithm

*ClosestInitPA* uses SVMs as its base learner. This fits well with many BioNLP tasks where SVMs deliver high performance (Giuliano et al., 2006; Lee et al., 2004). *ClosestInitPA* is based on the strategy of selecting the points which are closest to the current model's hyperplane (Tong and Koller, 2002) for human annotation. *ClosestInitPA* works best in situations with imbalanced data, which is often the case for BioNLP tasks. For example, in the AIMed dataset annotated with protein-protein interactions, the percentage of pairs of proteins in the same sentence that are annotated as interacting is only 17.6%.

SVMs (Vapnik, 1998) are learning systems that learn linear functions for classification. A statement of the optimization problem solved by soft-margin SVMs that enables the use of asymmetric cost factors is the following:

Minimize: $\frac{1}{2}\|\vec{w}\|^2 + C_+ \sum_{i:y_i=+1} \xi_i + C_- \sum_{j:y_j=-1} \xi_j$ (1)

Subject to: $\forall k : y_k[\vec{w} \cdot \vec{x}_k + b] \geq 1 - \xi_k$ (2)

where $(\vec{w}, b)$ represents the hyperplane that is learned, $\vec{x}_k$ is the feature vector for example k, $y_k$ in {+1,-1} is the label for example k, $\xi_k = \max(0, 1 - y_k[\vec{w} \cdot \vec{x}_k + b])$ is the slack variable for example k, and $C_+$ and $C_-$ are user-defined cost factors that trade off separating the data with a large margin and misclassifying training examples.

Let $PA = C_+/C_-$. PA stands for "positive amplification." We use this term because as the PA is increased, the importance of positive examples is amplified. *ClosestInitPA* is described in Figure 3. We have previously shown that setting PA based on a small initial set of data outperforms the more obvious approach of using the current labeled data to estimate PA.

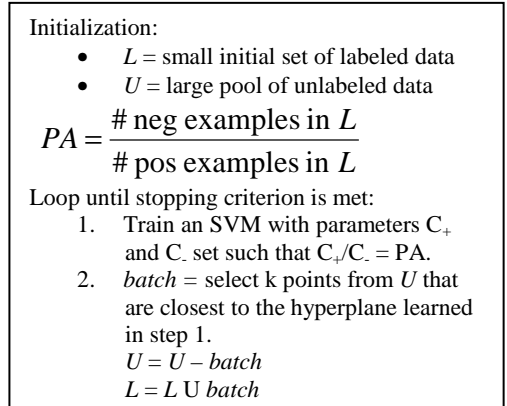

Figure 3. *ClosestInitPA* algorithm.

We have previously developed a stopping criterion called *staticPredictions* that is based on stopping when we detect that the predictions of our models on some unlabeled data have stabilized. All of the automatic stopping points in our results are determined using *staticPredictions*.

## 3 Experiments

**Protein-Protein Interaction Extraction**: We used the AImed corpus, which was previously used for training protein interaction extraction systems in (Giuliano et al., 2006). We cast RE as a binary classification task as in (Giuliano et al., 2006).

We do 10-fold cross validation and use what is referred to in (Giuliano et al., 2006) as the $K_{GC}$ kernel with $SVM^{light}$ (Joachims, 1999) in our experiments. Table 1 reports the results.

| StoppingPoint | Average # Labels | F Measure | |
|---|---|---|---|
| | | Random | AL |
| 20% | 1012 | 48.33 | 54.34 |
| 30% | 1516 | 49.76 | 54.52 |
| 40% | 2022 | 53.11 | 56.39 |
| 100% | 5060 | 57.54 | 57.54 |
| AutoStopPoint | 1562 | 51.25 | 55.34 |

Table 1. AImed Stopping Point Performance. "AutoStopPoint" is when the stopping criterion says to stop.

**Medline Text Classification**: We use the Ohsumed corpus (Hersh, 1994) and a linear kernel with $SVM^{light}$ with binary features for each word that occurs in the training data at least three times. Results for the five largest categories for one versus the rest classification are in Table 2.

| StoppingPoint | Average # Labels | F Measure | |
|---|---|---|---|
| | | Random | AL |
| 20% | 1260 | 49.99 | 61.49 |
| 30% | 1880 | 54.18 | 62.72 |
| 40% | 2500 | 57.46 | 63.75 |
| 100% | 6260 | 65.75 | 65.75 |
| AutoStopPoint | 1204 | 47.06 | 60.73 |

Table 2. Ohsumed stopping point performance. "AutoStopPoint" is when the stopping criterion says to stop.

**GENIA NER**: We assume a two-phase model (Lee et al., 2004) where boundary identification of named entities is performed in the first phase and the entities are classified in the second phase. As in the semantic classification evaluation of (Lee et al., 2004), we assume that boundary identification has been performed. We use features based on those from (Lee et al., 2004), a one versus the rest setup and 10-fold cross validation. Tables 3-5 show the results for the three most common types in GENIA.

| StoppingPoint | Average # Labels | F Measure | |
|---|---|---|---|
| | | Random | AL |
| 20% | 13440 | 86.78 | 90.16 |
| 30% | 20120 | 87.81 | 90.27 |
| 40% | 26900 | 88.55 | 90.32 |
| 100% | 67220 | 90.28 | 90.28 |
| AutoStopPoint | 8720 | 85.41 | 89.24 |

Table 3. Protein stopping points performance. "AutoStopPoint" is when the stopping criterion says to stop.

| StoppingPoint | Average # Labels | F Measure | |
|---|---|---|---|
| | | Random | AL |
| 20% | 13440 | 79.85 | 82.06 |
| 30% | 20120 | 80.40 | 81.98 |
| 40% | 26900 | 80.85 | 81.84 |
| 100% | 67220 | 81.68 | 81.68 |
| AutoStopPoint | 7060 | 78.35 | 82.29 |

Table 4. DNA stopping points performance. "AutoStopPoint" is when the stopping criterion says to stop.

| StoppingPoint | Average # Labels | F Measure | |
|---|---|---|---|
| | | Random | AL |
| 20% | 13440 | 84.01 | 86.76 |
| 30% | 20120 | 84.62 | 86.63 |
| 40% | 26900 | 85.25 | 86.45 |
| 100% | 67220 | 86.08 | 86.08 |
| AutoStopPoint | 4200 | 81.32 | 86.31 |

Table 5. Cell Type stopping points performance. "AutoStopPoint" is when the stopping criterion says to stop.

## 4 Conclusions

*ClosestInitPA* is well suited to many BioNLP tasks. In experiments, the annotation savings are practically significant for extracting protein-protein interactions, classifying Medline text, and performing biomedical named entity recognition.